%% file: paper.tex
\pdfoutput=1

\documentclass[11pt]{article}

\usepackage{acl}

\usepackage{times}
\usepackage{latexsym}

\usepackage[T1]{fontenc}

\usepackage[utf8]{inputenc}

\usepackage{microtype}

\usepackage{inconsolata}
\usepackage{pifont}

\usepackage{amsmath}
\usepackage{amsfonts}
\usepackage{amssymb}

\usepackage{booktabs}
\usepackage{multirow}

\usepackage{paralist}

\usepackage{graphicx}
\usepackage{enumitem}

\usepackage{color}

\usepackage[ruled,linesnumbered]{algorithm2e}

\usepackage{xspace}

\usepackage{subcaption}

\usepackage{float}

\newcommand{\cD}{\mathcal{D}}

\newcommand{\method}{WACO\xspace}

\title{\method: Word-Aligned Contrastive Learning for Speech Translation}

\author{
    Siqi Ouyang\textsuperscript{1}, Rong Ye\textsuperscript{2}, Lei Li\textsuperscript{1} \\ 
    \textsuperscript{1}University of California, Santa Barbara, CA, USA \\
    \texttt{siqiouyang@ucsb.edu, leili@cs.ucsb.edu} \\
    \textsuperscript{2}ByteDance AI Lab, Shanghai, China \\
    \texttt{yerong@bytedance.com}
}

\begin{document}
\maketitle

\begin{abstract}

\input{000abstract}
\end{abstract}

\section{Introduction}
\label{sec:intro}
\input{010intro}

\section{Related Work}
\label{sec:related}
\input{020related}

\section{Proposed Method: \method}
\label{sec:approach}
\input{030method}
\section{Experiments}
\label{sec:exps}
\input{040exp}

\section{Analysis}
\label{sec:analysis}
\input{050analysis}

\section{Conclusion}
\label{sec:conclusion}
\input{060conclusion}

\newpage
\section*{Limitations}
\label{sec:limit}
\input{070limitation}

\section*{Ethics Statement}
\label{sec:impact}
\input{080impact}

\section*{Acknowledgement}
Siqi Ouyang is supported by UCSB-IEE-Meta Collaborative Research Grant on AI.

\bibliography{paper}
\bibliographystyle{acl_natbib}

\appendix

\section{Appendix}
\label{sec:appendix}
\input{090appendix}

\end{document}

%% file: 000abstract.tex
End-to-end Speech Translation (E2E ST) aims to directly translate source speech into target text. 
Existing ST methods perform poorly when only extremely small speech-text data are available for training. 
We observe that an ST model's performance closely correlates with its embedding similarity between speech and source transcript. 
In this paper, we propose \textbf{W}ord-\textbf{A}ligned \textbf{CO}ntrastive learning (\textbf{\method}), a simple and effective method for extremely low-resource speech-to-text translation.
Our key idea is bridging word-level representations for both speech and text modalities via contrastive learning. 
We evaluate \method and other methods on the MuST-C dataset, a widely used ST benchmark, and on a low-resource direction Maltese-English from IWSLT 2023. Our experiments demonstrate that \method outperforms the best baseline by 9+ BLEU points with only 1-hour parallel ST data. Code is available at \url{https://github.com/owaski/WACO}.

%% file: 010intro.tex

End-to-end speech translation (E2E ST) directly translates speech in a source language to text in a target language, without intermediate pipelines. E2E ST has witnessed significant progress in translation quality 
\cite{duong-etal-2016-attentional,weiss17_interspeech,10.1109/ICASSP.2018.8461690,8683343,gaido-etal-2020-end,dong2021consecutive,ye2022cross}
. However, existing E2E ST methods still perform poorly when only a limited amount of parallel ST data are available. How can we build a highly performant ST model with extremely low resource, e.g. only 1 hour of parallel data (approximately a few hundred utterances)?

\begin{figure}[t]
    \begin{subfigure}[b]{0.48\linewidth}
        \centering
        \includegraphics[height=1.45in]{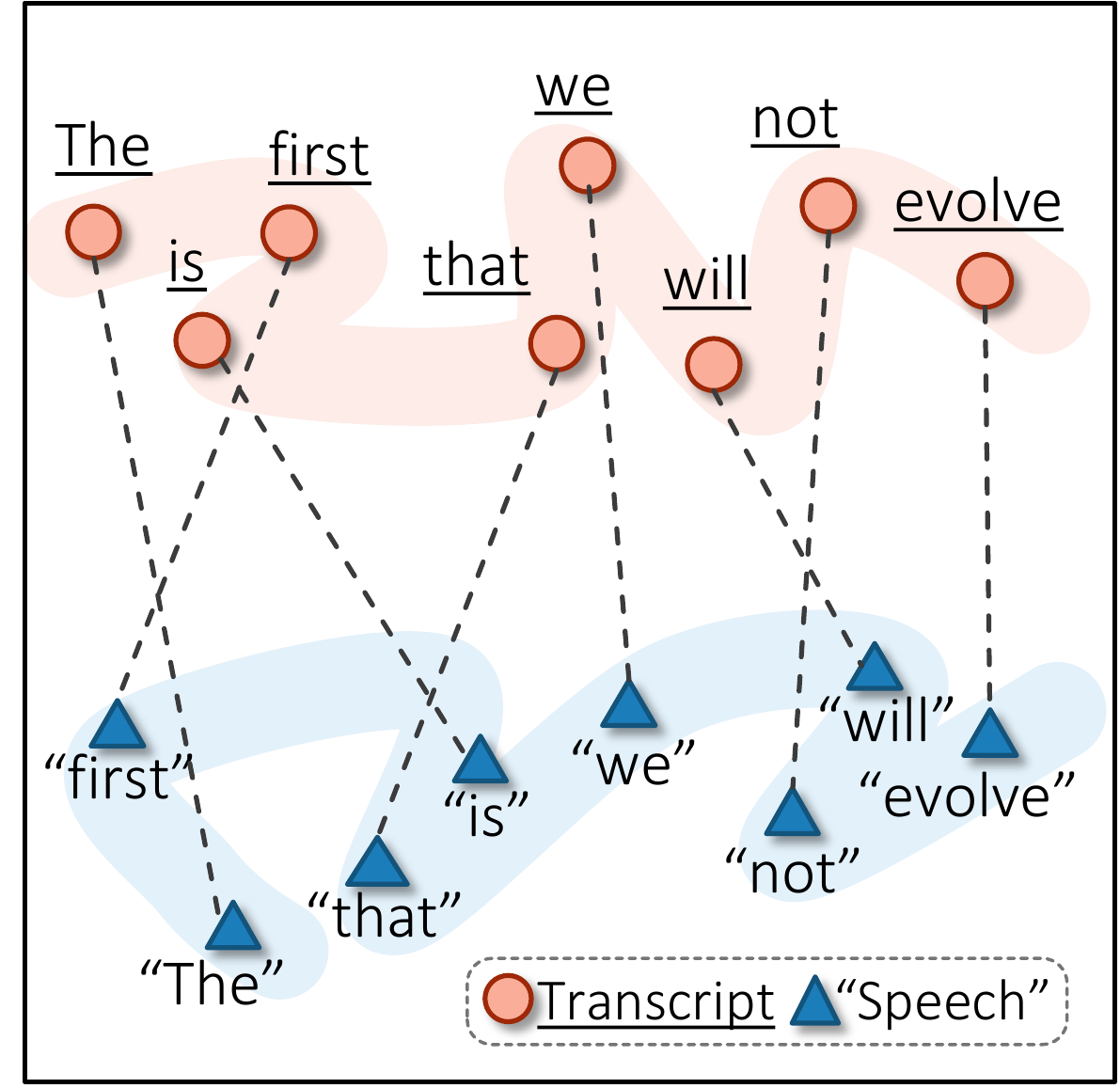}
        \caption{Existing \label{fig:current_emb}}
    \end{subfigure}
    \begin{subfigure}[b]{0.48\linewidth}
        \centering
        \includegraphics[height=1.45in]{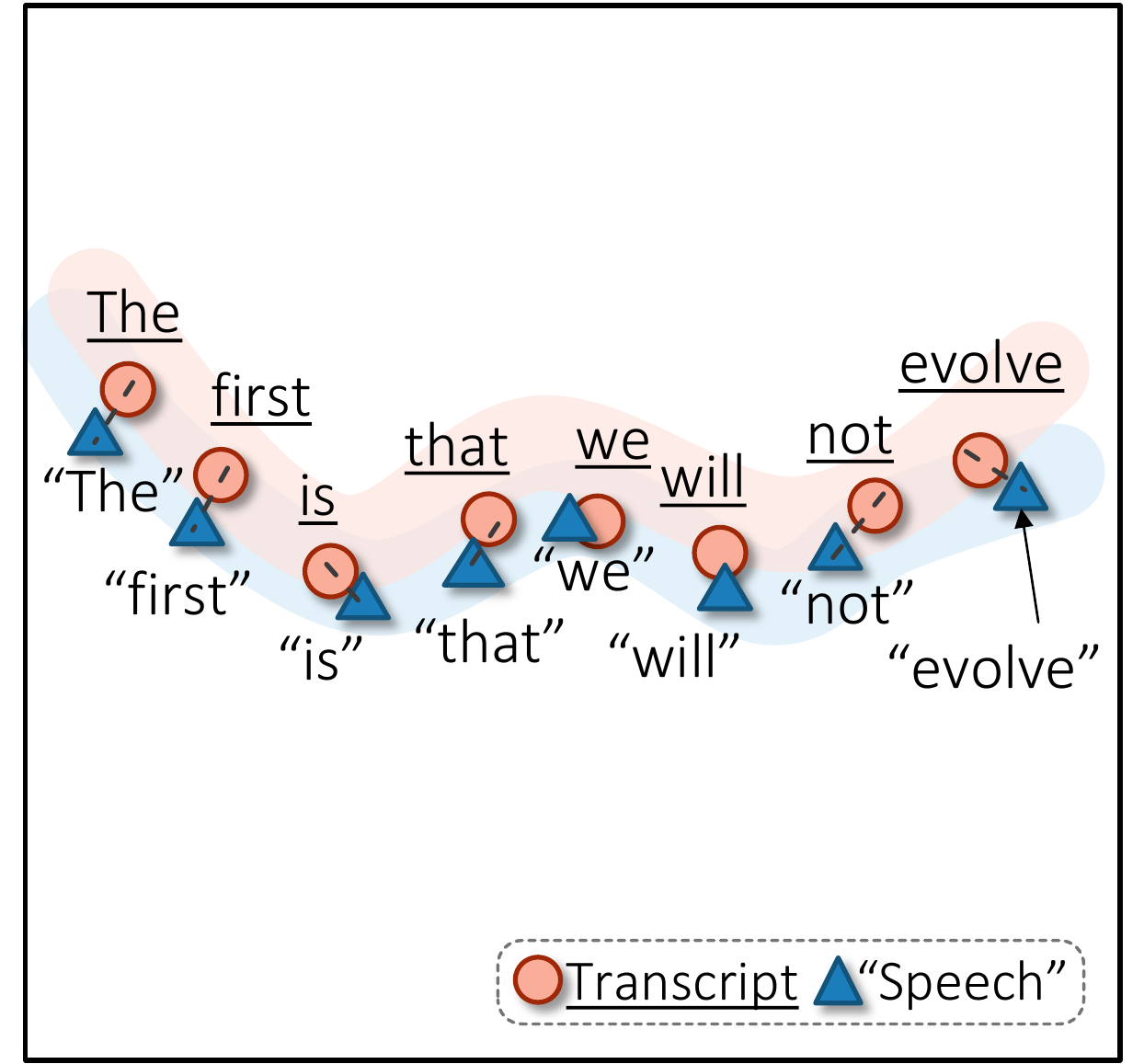}
        \caption{Ideal \label{fig:ideal_emb}}
    \end{subfigure}
    \caption{Better representations of speech and transcript text (projected to 2D) will lead to improved speech translation. 
    Ideal representations (\subref{fig:ideal_emb}) should not only be similar at the sequence level, but also closely align corresponding words between speech and transcript.}
    \label{fig:intro_strep}
    \vspace*{-0.5cm}
\end{figure}

We analyze the encoder representations from a directly learned ST model\footnote{We use XSTNet~\cite{ye2021end} as an example.} and find that its average embeddings from speech and transcript are similar at the sequence level but still not aligned well on the word level (Figure~\ref{fig:current_emb}). 
An ideal ST model should encode a speech utterance closely aligned with the representations of corresponding words in its transcript text (Figure~\ref{fig:ideal_emb}).
Prior methods attempt to use additional parallel data from machine translation (MT) and automatic speech recognition (ASR) to align speech and text representations. 
However, most do not explicitly reduce the word level representation gap among speech and text. 
We hypothesize that such misalignment at the word level between corresponding speech and transcript text is a critical cause of the inferior ST performance.

%

\begin{figure}[t]
    \centering
    \includegraphics[width=0.8\linewidth]{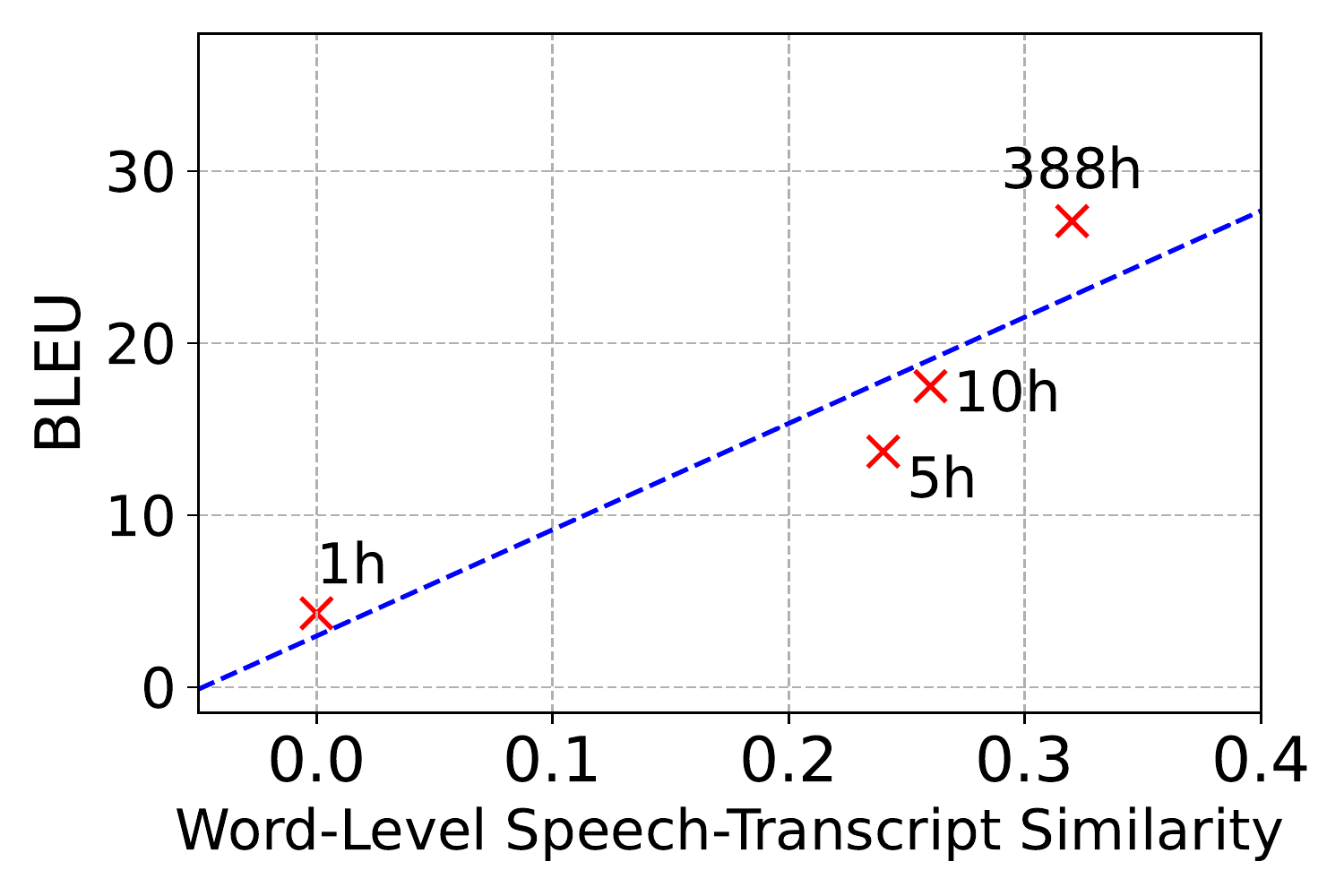}
    \caption{BLEU score of ST models trained on varying amounts of ST data (1/5/10/388 hours) and their cosine similarity scores between speech and transcript word embeddings. The ST performance closely correlates with speech-text representation similarity and both drop significantly when ST data is extremely limited.}
    \label{fig:bleu_data}
    \vspace*{-0.3cm}    
\end{figure}

We further observe such a misalignment phenomenon is severe when ST data is extremely low. 
We did a pilot study by training direct ST models using different sizes of ST data (1/5/10/388 hours) .
As shown in Figure \ref{fig:bleu_data}, we find that the translation performance highly correlates with the word-level embedding similarity between speech and transcript text. 
With fewer parallel ST data, the cross-modal similarity drops simultaneously with the BLEU score and almost reaches 0 given 1-hour ST training data. 
This observation suggests that the model can map both modalities into a partially aligned semantic space given sufficiently large ST data but fails when ST data is extremely small.


In this paper, we propose \method, a word-level contrastive learning method for extremely low-resource speech-to-text translation. 
Its key idea is to promote representation similarity among corresponding words in speech and text and to push non-corresponding representations further apart in embedding spaces. 
Furthermore, it can use additional large ASR data to improve word-level representation learning. 
We conduct experiments on the MuST-C benchmark with varying data sizes and a real-world low-resource direction Maltese-English (Mt-En) from IWSLT 2023 low-resource track. 
We also analyze the word-level representation similarity from the learned models. 
Our contributions are:
\begin{itemize}[itemsep=1pt, leftmargin=10pt, parsep=0pt, topsep=1pt]
    \item We propose a new method for speech translation, \method, which explicitly aligns speech-transcript representations of corresponding words.
    \item We verify the effectiveness of \method on MuST-C and IWSLT Mt-En. \method outperforms previous strong methods by 2.0-9.8 BLEU points with only 1 and 10 hours of parallel ST data.
    \item We further demonstrate that that \method indeed learns a better-aligned representation of speech and text at the word level which correlates well with its ST performance. 
    
\end{itemize}

%% file: 020related.tex

\noindent\textbf{End-to-end ST}~
Due to error propagation and high latency in cascaded ST systems, \citet{berard2016listen,duong-etal-2016-attentional} first proposed to translate source speech into target text directly without generating the intermediate transcript. The major difficulty in training end-to-end ST systems is the lack of direct ST data. Though many ST datasets~\cite{wang21s_interspeech,cattoni2021must} were proposed in recent years, the amount of ST data is still much less than that of MT and ASR. To overcome the data scarcity problem, methods including data augmentation \cite{park2019specaugment,mccarthy2020skinaugment,lam-etal-2022-sample,10.1016/j.neunet.2022.01.016}, self-training \cite{pino2020self}, multi-tasking \cite{le2020dual,tang2021general,tang-etal-2021-improving,ye2021end,pmlr-v162-zhang22i} and pre-training \cite{berard2018end,bansal2019pre,wu2020self,wang2020curriculum,alinejad2020effectively,dong2021consecutive,zheng2021fused,slam,ao-etal-2022-speecht5,tang-etal-2022-unified} have been proposed. \method is a novel approach that can be applied in existing multi-tasking and pre-training frameworks to improve ST performance. 


\noindent\textbf{Cross-modal representation learning}~
Researchers realized recently that the misalignment between speech and text representation hinders the knowledge transfer from external data \cite{liu2020bridging,LUT, xu2021stacked,han2021learning,ye-etal-2022-cross,npdae2est,zeroshotST}. \citet{liu2020bridging} shrank the speech representation to match the length of text representation and also closed the representational gap by minimizing their L2 distance. \citet{xu2021stacked} mapped speech representation to text representation through both the Connectionist Temporal Classification (CTC)~\cite{graves2006connectionist} distribution and a mapping layer. \citet{LUT} proposed a cross attention layer to force the speech-text alignment. \citet{han2021learning} developed a novel architecture enabling fixed-length shared semantic space for both modalities. \citet{saxon21_interspeech} proposed a hierarchical speech understanding system leveraging both ASR and text understanding data. \citet{ye-etal-2022-cross} employed sentence-level contrastive loss to reduce the modality gap and ach`ieved state-of-the-art results on MuST-C. Our method, however, works on word-level instead of sentence-level and empirically provides both better performance and higher data efficiency. \citet{fang-etal-2022-stemm} also proposes to align the word-level representations between speech and text, but their method heavily relies on target translation while our method only requires ASR data for modality gap reduction.



%% file: 030method.tex
In this section, we describe our approach to develop effective speech translation models with extremely low-resource parallel data.


\subsection{Problem Formulation}
\label{sec:formulation}

A typical ST corpus $\mathcal{D}^\text{ST}$ contains speech $s$ and its transcript $x$ in a source language and translation text $y$ in a target language. Equivalently, $\mathcal{D}^\text{ST} = \{(s,x,y)\}$ and ASR corpus can be similarly defined as $\mathcal{D}^\text{ASR} = \{(s,x)\}$.

Given $\cD^\text{ST}$ and $\cD^\text{ASR}$ as training sets, the E2E ST model needs to translate speech $s$ into translation $y$ accurately without generating transcript $x$ in the intermediate steps. There are two settings:
\begin{itemize}[itemsep=1pt, leftmargin=10pt, parsep=0pt, topsep=1pt]
    \item \textbf{Regular ST}: Training includes large ST triplet data. In this paper, we regard the entire MuST-C training set as the regular setting ($|\cD^\text{ST}|\approx$400 hours).
    \item \textbf{Low-Resource ST}: Training has very limited ST data but plenty of ASR data, i.e., $|\cD^\text{ST}| \ll |\cD^\text{ASR}|$. In this paper, ST data below 10 hours is regarded as low-resource ST. Many African and native American languages belong to this setting.
\end{itemize}

In this paper, we focus on low-resource ST. In addition, we also include external MT data for both settings. The size of MT dataset is much larger than ST dataset.


\begin{figure}[t]
    \setlength{\belowcaptionskip}{-0.4cm}
    \centering
    \includegraphics[width=0.8\linewidth]{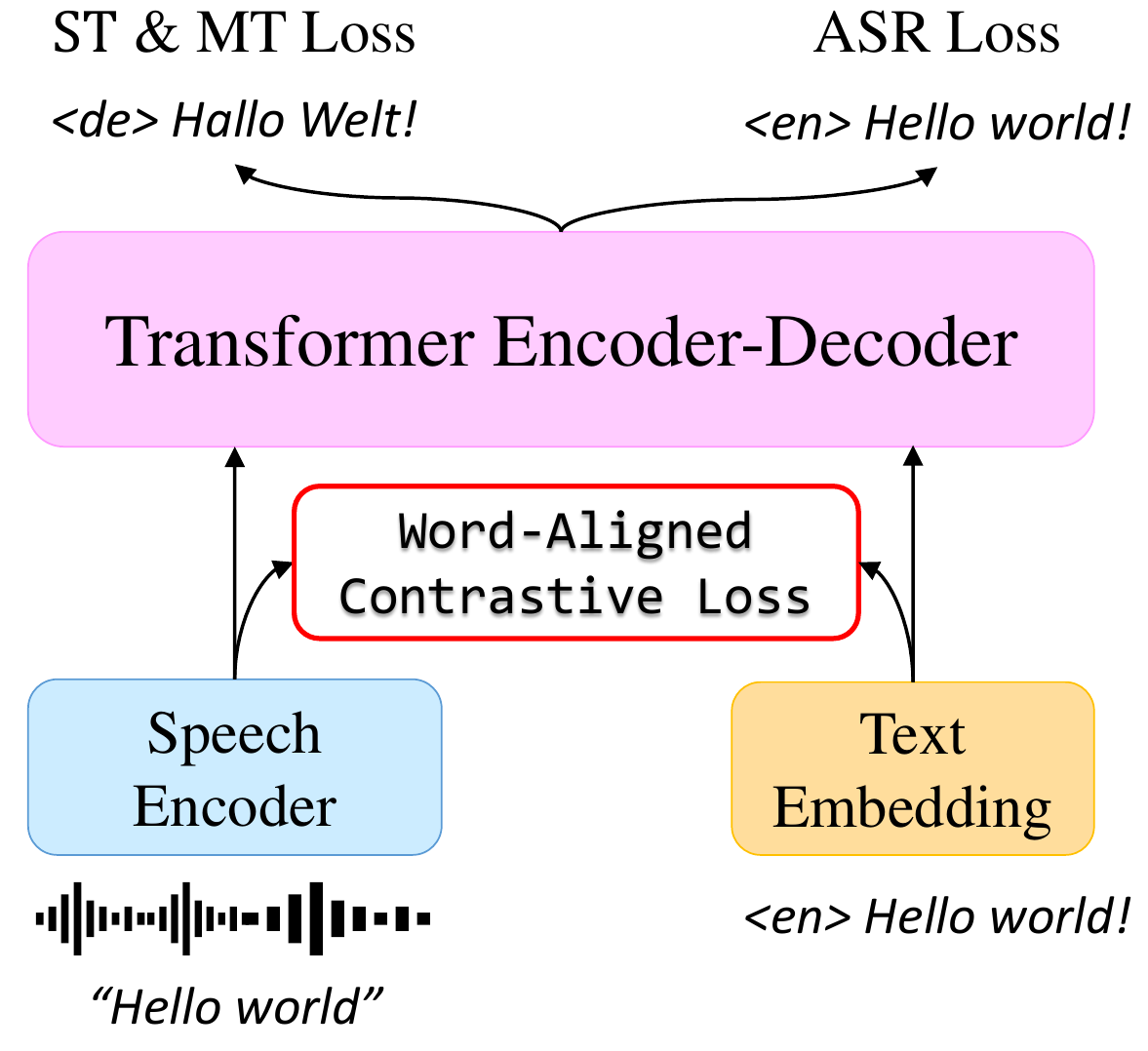}
    \caption{Model architecture of \method. During training, it accepts either speech or text input and outputs text sequence. We introduce word-aligned contrastive loss to align  representations of speech and transcript text. At inference time, it only needs speech input.}
    \label{fig:arch}
\end{figure}

\begin{figure*}[ht]
    \setlength{\abovecaptionskip}{-0.cm}
    \setlength{\belowcaptionskip}{-0.4cm}
    \centering
    \includegraphics[width=\textwidth]{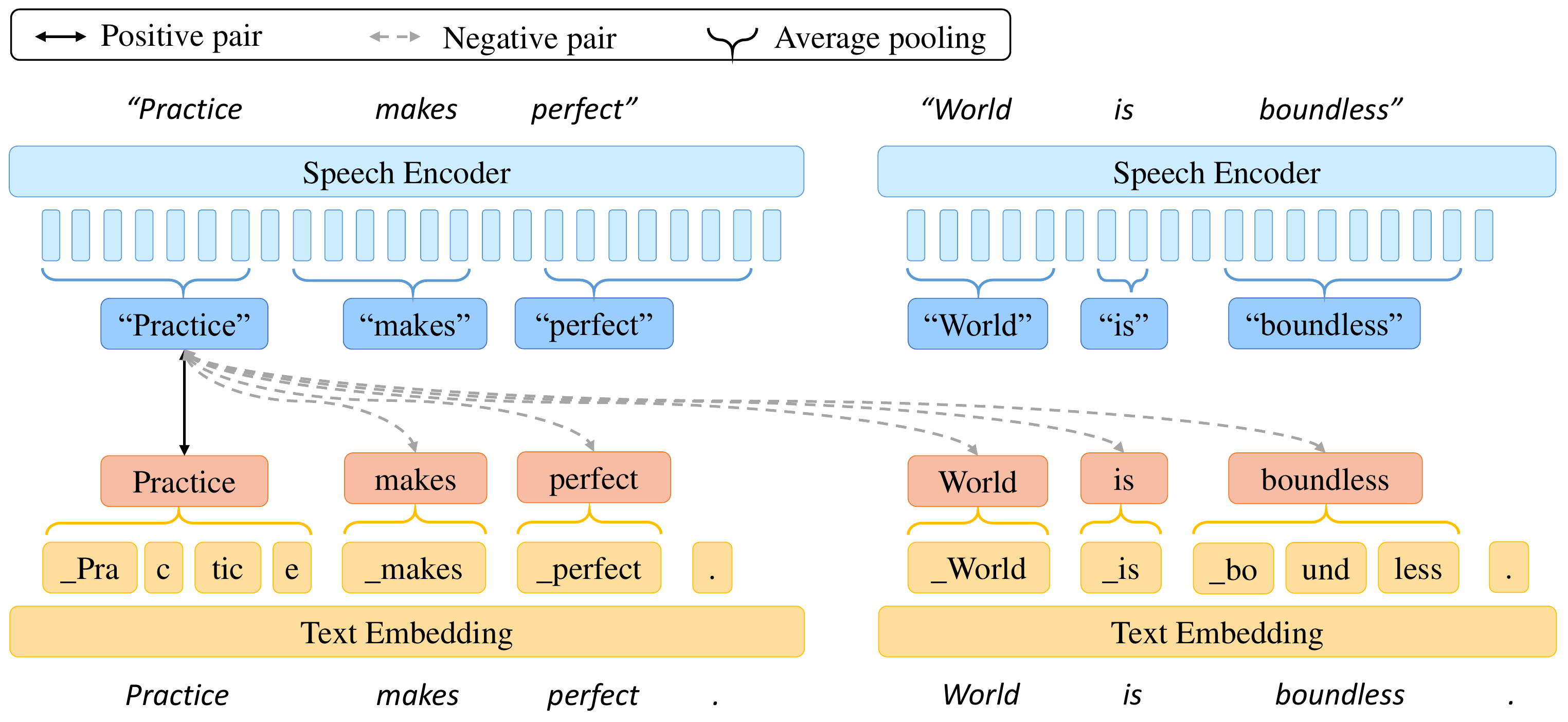}
    \caption{An illustration of word-aligned contrastive learning for a batch of two data points. Speech and text are passed through speech encoder and text embedding respectively to obtain embeddings. Then we group embeddings by word-level average pooling for both modalities. Average speech and text embeddings for the same word are treated as the positive pair and average embeddings for different words are treated as the negative pairs.}
    \label{fig:waco}
\end{figure*}

\subsection{Model Architecture}
\label{sec:arch}

Figure~\ref{fig:arch} illustrates \method model architecture. \method consists of 3 modules: a speech encoder, a text embedding layer and a joint Transformer encoder-decoder. During training, it inputs either speech or text sequence and outputs text sequence. At inference time, the input is only speech. 

\noindent\textbf{Speech Encoder} extracts contextualized acoustic embeddings from the raw waveform. It consists of wav2vec 2.0 \cite{baevski2020wav2vec} and 2 downsampling layers. Wav2vec 2.0 is one of the state-of-the-art self-supervised models pre-trained on unlabeled English speech corpus. It produces contextualized speech embeddings. It consists of 7 convolutional layers as the audio feature extractor and a 12-layer Transformer as the contextual encoder. On top of wav2vec 2.0, we further add 2 convolutional layers with stride of 2 to downsample the embedding sequence. It reduces the length and alleviates the length discrepancy between speech and text embeddings.

\noindent\textbf{Text Embedding} is a lookup table that maps text tokens into embeddings. 

\noindent\textbf{Transformer Encoder-Decoder} accepts outputs from either both the speech encoder or the text embedding layer. The configuration is the same as the 6-layer vanilla Transformer \cite{vaswani2017attention}. Specifically, the encoder further extracts contextualized high-level semantic features from both modalities and the decoder generates a token sequence for different tasks. Besides, since we are using general Transformer architecture, both the text embedding layer and the Transformer can be pre-trained on additional MT data.

\subsection{Word-Aligned Contrastive Learning}
\label{sec:loss}

To alleviate the misalignment between speech and transcript representations, we propose word-aligned contrastive learning to bring speech and text embeddings closer in a fine-grained level (Figure \ref{fig:waco}).

Given a speech-transcript pair $(s, x)$. The transcript is tokenized by a Byte-Pair-Encoding (BPE) tokenizer into a sequence of BPE tokens $x=(x_1,x_2,\cdots,x_n)$. Though BPE is the default option for text tokenization, whole words preserves acoustic boundary better than BPE tokens. Thus, we group $n$ BPE tokens back into $m$ whole words where $w_i=x[l_i^t:r_i^t]$ for $i=1,2,\cdots,m$ where superscript $t$ stands for text feature and $l_i,r_i$ refer to the range of BPE tokens that word $i$ is tokenized into. In the example of Figure \ref{fig:waco}, $x =$ (\_Pra, c, tic, e, \_makes, \_perfect, .), and $w_1 = $ (\_Pra, c, tic, e), $w_2=$ (\_makes), $w_3=$ (\_perfect), and $w_4$ = (.). Then we align whole words $w_1,w_2,\cdots,w_m$ with speech $s=(s_1,s_2,\cdots,s_{|s|})$ by a forced aligner. Here we use Montreal Forced Aligner\footnote{ \url{https://github.com/MontrealCorpusTools/Montreal-Forced-Aligner}} (MFA). This provides us time interval $1\leq l_i^s \leq r_i^s\leq |s|$ for word $w_i$ where superscript $s$ stands for speech feature. 

Now we have identified $m$ corresponding pairs of speech segments $s[l_i^s:r_i^s]$ and words $x[l_i^t : r_i^t]$ for $i=1,2,\cdots,m$. The representations of them are obtained as follows,
\begin{align}
    f_i^s &= \text{MeanPool}(\text{S-Enc}(s)[\tilde{l_i^{s}} : \tilde{r_i^s}]) \label{eq:f_s}\\
    f_i^t &= \text{MeanPool}(\text{T-Emb}(x)[l_i^t : r_i^t])
    \label{eq:f_t}
\end{align}
where S-Enc is the speech encoder, T-Emb is the text embedding layer,~$\tilde{l_i^s} = \frac{l_i^s}{|s|}|\text{S-Enc}(s)|$ and~$\tilde{r_i^s} = \frac{r_i^s}{|s|}|\text{S-Enc}(s)|$ refer to the relative indices given the audio representation length shrinkage after the speech encoder.

We treat $f_i^s$ and $f_i^t$ as a positive pair and treat $f_i^s$ and other words in the same batch as negative pairs and we apply multi-class N-pair contrastive loss \cite{NIPS2016_6b180037} on them:
\begin{flalign}
    \ell_\text{CTR}(\mathcal{B}) &= \nonumber \\ 
    -\mathop{\mathbb{E}}_{f^s_i,f^t_i\in \mathcal{B}}&\left[\log \frac{\exp(sim(f^s_i,f^t_i)/\tau)}{\sum_{f^t_{j\neq i}\in \mathcal{B}}\exp(sim(f_i^s,f_j^t)/\tau)}\right] \label{eq:l_ctr}
\end{flalign}
where $\mathcal{B}$ is the current batch, $\tau$ is the temperature hyper-parameter, $sim()$ is used to measure the distance between two representations, and we use cosine similarity $sim(a,b)=a^\top b/\|a\|\|b\|$.



\subsection{Training and Inference}
\label{sec:training_strategy}

\paragraph{Cross-Modal Pre-training} We first pre-train text embedding and joint Transformer on external MT dataset (e.g., WMT dataset). 
Then we train the MFA model on ASR data $\cD^\text{ASR}$ to obtain alignments, and further train our ST model using word-aligned contrastive loss on $\cD^\text{ASR}$
\begin{align}
    \mathcal{L}^\text{PT} = \mathop{\mathbb{E}}_{\mathcal{B}\subseteq \cD^\text{ASR}} \left[\ell_\text{CTR}(\mathcal{B})\right],
\end{align}
where $\ell_\text{CTR}(\mathcal{B})$ is defined in Equation~\ref{eq:l_ctr}.
The pre-training stage aims to map speech and text embeddings into a shared semantic space using ASR and MT data. 

\paragraph{Multi-task Fine-tuning}

We fine-tune our model using the multi-task cross-entropy losses, as suggested in \citet{ye2021end}, and contrastive loss.
\begin{align}
    \mathcal{L}^\text{FT} = \mathcal{L}_\text{CE} + \lambda\mathcal{L}_\text{CTR}
\end{align}
where
\begin{align}
    \mathcal{L}_\text{CE} &= \mathop{\mathbb{E}}_{(s,x,y)\in\cD^\text{ST}}\left[\ell_\text{ST} + \ell_\text{MT} + \ell_\text{ASR} \right] \label{eq:l_ce} \\
    \
    \mathcal{L}_\text{CTR} &= \mathop{\mathbb{E}}_{\mathcal{B}\subseteq \cD^\text{ST}} \left[\ell_\text{CTR}(\mathcal{B})\right].
\end{align}

Cross entropy losses are derived directly from the triplet dataset $\cD^\text{ST}$, 
\begin{align}
    \ell_\text{ST}(s,y) &= -\log P(y|s) \\
    \ell_\text{MT}(x,y) &= -\log P(y|x) \\
    \ell_\text{ASR}(s,x) &= -\log P(x|s).
\end{align}
$\lambda$ is the hyper-parameter controlling the weight of contrastive loss. 

\paragraph{Inference}

During inference, the model accepts speech frames as input of speech encoder and decodes translation in the target language through beam search. No source transcript text is needed during inference. 

%% file: 040exp.tex



\subsection{Datasets}

\paragraph{MuST-C} We conduct experiments on the MuST-C dataset\footnote{Released under CC BY NC ND 4.0 International}\textsuperscript{,}\footnote{Here we refer to MuST-C v1.0.}~\cite{digangi2019must}, one of the largest ST benchmark datasets containing translations from English to 8 languages collected from TED Talks. Each language direction involves around 400 hours of audio recordings. Limited by computing resources, we examine our method on three language directions: English-German/French/Spanish (En-De/Fr/Es). 


\paragraph{MuST-C Low-Resource} To examine our method in the extremely low-resource settings, we manually create ASR and ST subsets from the MuST-C En-De training set. Specifically, we build 10-hour, 100-hour and 370-hour ASR subsets and 1-hour and 10-hour ST subsets respectively through random sampling. 

\paragraph{IWSLT Low-Resource} We also evaluate our method on Maltese to English translation in IWSLT 2023 low-resource track\footnote{\url{https://iwslt.org/2023/low-resource}}. We use the officially provided ST triplets as ST data and build the ASR dataset by combining the the audio-transcript part from the official ST data and CommonVoice~\cite{ardila-etal-2020-common}\footnote{We do not use the officially provided ASR dataset due to its severe audio-transcript misalignment.}. We remove silences and randomly partition the data. Finally we obtain 1 hour of ST triplets and 10 hours of ASR pairs as training and development set and 0.1 hour of ST triplets as test set.

\paragraph{External ASR} We also introduce LibriSpeech \cite{panayotov2015librispeech} as the external ASR dataset. LibriSpeech is the \textit{de facto} public English ASR benchmark\footnote{Released under CC BY 4.0} containing 960 hours of audiobook speech. We build a 1330-hour English ASR dataset by combining MuST-C and LibriSpeech.

\paragraph{External MT} Additionally, we introduce external WMT En-De/Fr/Es datasets \cite{bojar2016findings} for each language direction to pre-train text embedding and Transformer. 
We also introduce Flores-200~\cite{team2022NoLL}, a massively multilingual machine translation dataset, for Mt-En direction.
As shown in previous work \cite{xu2021stacked,ye2021end}, MT pre-training greatly improves ST performance. 

The statistics of datasets above are listed in Appendix~\ref{sec:data_stat}.

\subsection{Experimental Setups}
\label{sec:exp_setup}

\paragraph{Model Configurations} In MuST-C experiments, we use wav2vec 2.0 base model\footnote{\url{https://dl.fbaipublicfiles.com/fairseq/wav2vec/wav2vec_small.pt}} in our speech encoder which is solely pre-trained on 960-hour English audio. It consists of a 7-layer convolutional feature extractor and 12 Transformer encoder blocks with 768 hidden units. Two down-sampling convolutional layers have kernel size 5, stride size 2 and hidden size 512 or 1024 depending on the Transformer hidden size. For En-De/Fr/Es directions, the Transformer encoder-decoder has 6 encoder and decoder layers with hidden size 512, 2048 FFN hidden units and 8 attention heads. For Mt-En direction, it has 12 layers each with hidden size 1024, 4096 FFN hidden units and 16 attention heads. 

\paragraph{Preprocessing}  The input speech is the raw 16-bit 16kHz mono-channel waveform. We filter speech that is either too long (>480k frames) or too short (<1k frames) out. This results in 388/471/480 hours of speech being retained as ST training data for En-De/Fr/Es directions. We jointly tokenize the transcripts and translations for each language direction using SentencePiece \cite{kudo2018sentencepiece} with a vocabulary size set to 10k. To conduct forced alignment required by \method (see section \ref{sec:loss}), we first remove punctuations and group whole words by identifying special space token in the vocabulary. Then we use the MFA to train forced aligners on $\cD^\text{ASR}$ to align English speech and words. \textbf{The amount of ASR data used to train forced aligner is the same as that used for training \method.} Due to vocabulary mismatch between the MFA and our SentencePiece model, a small number of speeches and transcripts (e.g., 18h for En-De) cannot be aligned and we simply ignore them when doing contrastive learning.

\input{table/main_table}

\paragraph{Training} Transformer and text embedding are pre-trained on the external WMT dataset for En-De/Fr/Es (MT training details can be found in Appendix \ref{sec:mt_pt}). For Mt-En direction, we directly initializes the Transformer and text embedding with the NLLB-600M model~\cite{team2022NoLL}\footnote{\url{https://github.com/facebookresearch/fairseq/tree/nllb}} pre-trained on Flores-200. For both cross-modal pre-training and multi-task fine-tuning, we set contrastive temperature $\tau=0.2$ and optimize our model by Adam optimizer \cite{kingma2015adam} ($\beta_1=0.9,\beta_2=0.98$) with learning rate 1e-4 and 25k warm-up steps. After the warm-up, the learning rate is decayed following the inverse square root schedule. The effective batch size is 16 million frames. We set dropout rate to 0.1. For pre-training, we save the checkpoints with the best contrastive loss on the validation set. For fine-tuning, we save the checkpoints with the best BLEU on the validation set and average the last 10 saved checkpoints. Also, we set label smoothing to 0.1 for the cross-entropy losses, $\lambda=0$ in low-resource ST and $\lambda=1$ in ST with full data. All models are trained on Nvidia A6000 GPUs.

\paragraph{Inference and Evaluation} During inference, we run beam search with beam size 10 and length penalty 0.6/1.0/0.1/0.3 for En-De/Fr/Es and Mt-En directions respectively. For evaluation, we report case-sensitive detokenized BLEU scores on MuST-C tst-COMMON and IWSLT Mt-En test set using sacreBLEU \cite{post2018call}\footnote{BLEU signature: nrefs:1|bs:1000|seed:12345|case:mixed|
eff:no|tok:13a| smooth:exp|version:2.0.0}.

\paragraph{Baselines} In low-resource ST settings, we compare our method with three baselines: 
\begin{itemize}[itemsep=1pt, leftmargin=10pt, parsep=0pt, topsep=1pt]
\item \textbf{Base}: This baseline ignores $\cD^\text{ASR}$ and only optimizes cross entropy loss in Equation~\ref{eq:l_ce} on $\cD^\text{ST}$.
\item \textbf{Base+CTC}: This baseline, on top of \textbf{Base}, applies CTC loss on $\cD^\text{ASR}$ to align speech and text representations. In particular, we add a linear layer after the speech encoder to predict the text BPE token at each frame and fix its weight with text embedding. We only include CTC with BPE tokenization here since it performs consistently better than its phoneme counterpart (details in Section~\ref{sec:ctc_bpe}).
\item \textbf{ConST}: This baseline adds a coarse-grained contrastive loss on $\cD^\text{ASR}$ on top of \textbf{Base} to reduce modality gap as in~\citet{ye-etal-2022-cross}. Instead of word-level alignment, \textbf{ConST} conducts contrastive learning on sentence-level average speech and text embeddings. Hyper-parameters are directly borrowed from \citet{ye-etal-2022-cross}.
\end{itemize}

In regular ST with full MuST-C data, we compare our method with other existing works.

\subsection{Main Results}

\paragraph{Low-Resource ST} Results are shown in Table \ref{tab:main_table}. The ASR data for cross-modal pre-training varies from 10 hours to 1330 hours, and the ST data for multi-task fine-tuning varies from 1 hour to 10 hours. \textbf{\method} consistently outperforms baseline methods in all data configurations and language directions. In particular, our model achieves a BLEU score of 14.1 for En-De and 13.3 for Mt-En with only 1h ST and 10h ASR data and 21.0 for En-De with only 10h ST and 100h ASR data. With 1330h ASR data, \method even pushes the BLEU score to 17.5 and 22.9. More surprisingly, we find that \textbf{\method} has a further advantage when using less ASR data. When reducing ASR data from 370 hours to 100 hours, the BLEU score increases (\textbf{\method} vs \textbf{Base+CTC}, \textbf{ConST}) are enlarged from +2.0,+4.9 to +4.0,+8.9 in 1h ST setting. This demonstrates that \method is more efficient than the baseline methods especially in low-resource setting.

\input{table/full_mustc}

\paragraph{Regular ST} Results are shown in Table \ref{tab:full_data}. Here we are using the entire MuST-C training set as in previous works to enable fair comparison, which means $\cD^\text{ST}$ has full MuST-C training data. \method is competitive with previous models such as STEMM and ConST in all three language directions. Note that SpeechUT and STPT achieve the highest BLEU scores in all directions, but both SpeechUT and STPT leverages additional speech data (60k hours) or ASR data (100 hours) and employ a different model architecture.

%% file: table/main_table.tex
\begin{table*}[t]
\setlength{\belowcaptionskip}{-0.3cm}
\centering

\begin{tabular}{l|cccccccc} 
    \toprule

    \textbf{Direction} & \multicolumn{7}{c}{\textbf{En-De}} & \textbf{Mt-En} \\
    
    \cmidrule(lr){2-8} \cmidrule(lr){9-9}
    \textbf{ST Data} &  \multicolumn{4}{c}{1h} & \multicolumn{3}{c}{10h} & 1h \\
    \cmidrule(lr){2-5} \cmidrule(lr){6-8} \cmidrule(lr){9-9}
    \textbf{ASR Data} & 10h & 100h & 370h & 1330h & 100h & 370h & 1330h & 10h \\
    \midrule
    Base       & 4.3 & 4.3  & 4.3  & 4.3  & 17.5 & 17.5 & 17.5 & 2.2 \\
    Base+CTC   & 0.2 & 12.6 & 14.6 & 14.7 & 18.3 & 20.4 & 20.0 & 3.0 \\
    ConST      & 3.0 & 7.3  & 11.7 & 13.7 & 16.9 & 18.6 & 19.6 & 4.0 \\
    \method    & \textbf{14.1} & \textbf{16.2} & \textbf{16.6} & \textbf{17.5} & \textbf{21.0} & \textbf{22.7} & \textbf{22.9} & \textbf{13.3}  \\
    \bottomrule
\end{tabular}


\caption{Case-sensitive detokenized BLEU scores on MuST-C En-De \texttt{tst-COMMON} set and manually created IWSLT Mt-En test set of models pre-trained on ASR data using different cross-modal methods and fine-tuned on ST data. All models share the same W2V2-Transformer architecture. \textbf{Base} ignores ASR data and only fine-tunes on ST data, while other baselines pre-train on ASR data using \textbf{CTC}, sentence-level contrastive (\textbf{ConST}) and word-aligned contrastive (\textbf{WACO}) losses.}
\label{tab:main_table}
\end{table*}

%% file: table/full_mustc.tex
\begin{table}[t]
    \setlength{\belowcaptionskip}{-0.5cm}
    \centering
    \small
    \resizebox{\linewidth}{!}{
    \begin{tabular}{l|ccc} 
        \toprule
        \textbf{Models} & \textbf{En-De} & \textbf{En-Fr} & \textbf{En-Es} \\ \midrule
        W-Transf.~\cite{ye2021end} & 23.6 & 34.6 & 28.4 \\
        SpeechT5~\cite{ao-etal-2022-speecht5} & 25.2 & 35.3 & - \\
        FAT-ST~\cite{zheng2021fused} & 25.5 & - & 30.8 \\
        JT-S-MT~\cite{tang-etal-2021-improving} & 26.8 & 37.4 & 31.0 \\
        Chimera~\cite{han2021learning} & 27.1 & 35.6 & - \\
        XSTNet~\cite{ye2021end} & 27.8 & 38.0 & 30.8 \\
        SATE~\cite{xu2021stacked} & 28.1 & - & - \\
        STEMM~\cite{fang-etal-2022-stemm} & \textbf{28.7} & 37.4 & 31.0 \\
        ConST~\cite{ye-etal-2022-cross} & 28.3 & \textbf{38.3} & \textbf{32.0} \\
        \method & 28.1 & 38.1 & \textbf{32.0} \\ \midrule
        STPT~\cite{tang-etal-2022-unified}* & 29.2 & 39.7 & 33.1 \\
        SpeechUT~\cite{zhang2022speechut}* & 30.1
 & 41.4 & 33.6 \\\bottomrule
    \end{tabular}
    }
    \caption{Case-sensitive detokenized BLEU scores on MuST-C En-De \texttt{tst-COMMON} set of models trained on full MuST-C training set. *Note that SpeechUT and STPT leverage more speech (60k hours) or ASR data (100 hours).}
    \label{tab:full_data}
\end{table}

%% file: 050analysis.tex
\label{sec:analysis}


\begin{figure*}[t]
    \setlength{\abovecaptionskip}{-0.cm}
    \centering
    \includegraphics[width=\linewidth]{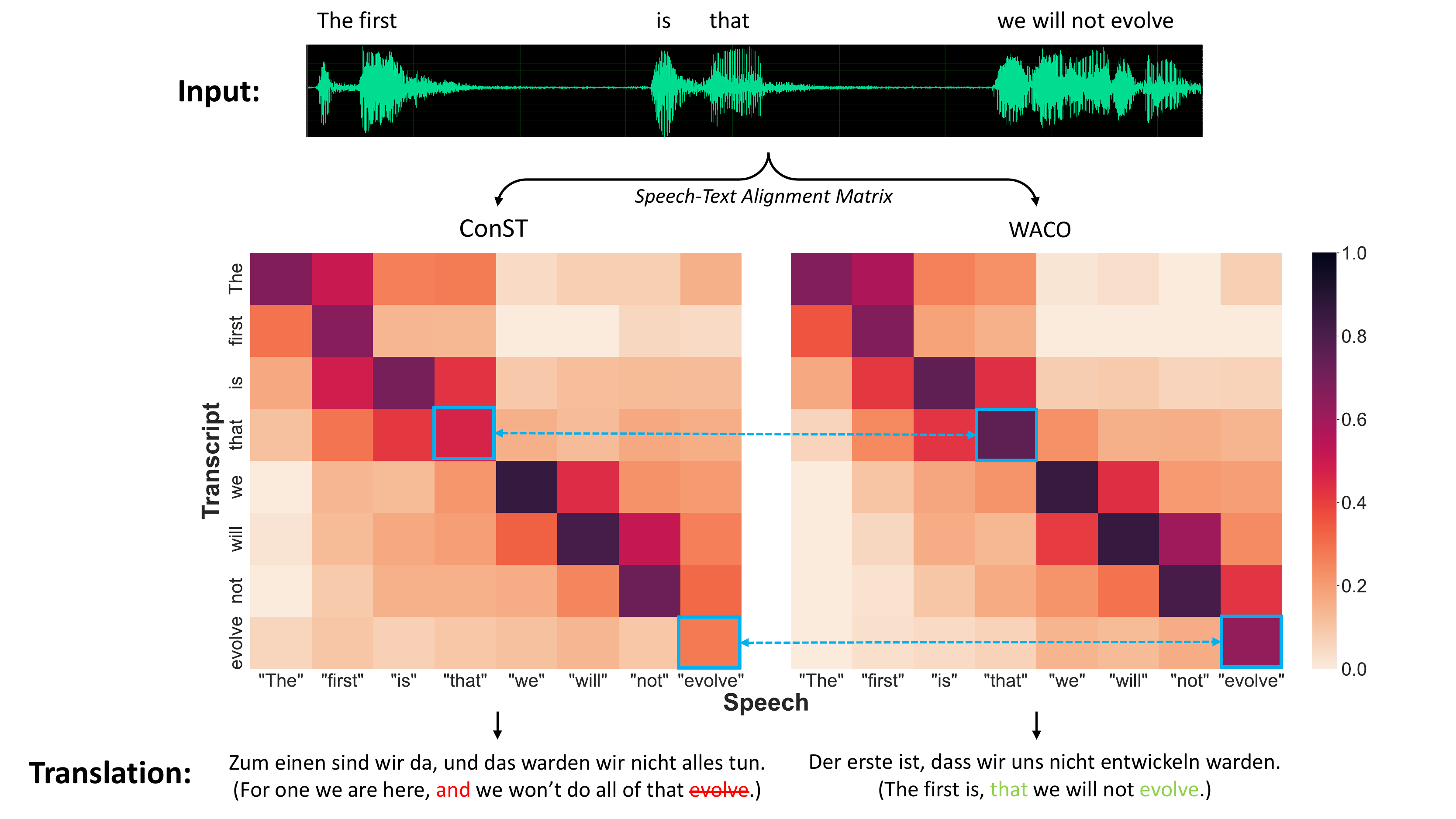}
    \caption{An example showing that \method can capture the word-level details better than ConST. The matrix illustrates pairwise cosine similarity between word-level average embeddings of speech and transcript. \method aligns two modalities well while ConST fails to align word ``that'' and ``evolve''. Though ConST still provides higher sentence-level similarity than \method (0.60 for ConST and 0.58 for \method), its translation is not as accurate as our method due to misaligned words.}
    \label{fig:case}
\end{figure*}

\subsection{Why is Word-level Contrastive Loss Better than Sentence-level Contrastive Loss?}
\label{sec:fine-grain}

Intuitively, only aligning the representations between speech and text at the sentence level cannot assure that model captures the accurate word correspondence between these two modalities. 

First, we measure the cosine similarity between speech embedding and transcript text embedding using models (ConST and WACO) pre-trained on 370h ASR dataset and fine-tuned on 1h ST dataset. The embeddings of speech and transcript text are calculated based on Equation~\ref{eq:f_s} and~\ref{eq:f_t} respectively.
The result is shown in Table~\ref{tab:avg_cos}. 
\method achieves more accurate word-level alignment than ConST (0.65 \textit{v.s.} 0.44), which indicates \textbf{\method can handle word-level details inside a sentence better.} Besides, though not optimized for sentence-level alignment, \method still achieves close sentence-level similarity with ConST (0.30 \textit{v.s.} 0.33).

We also show an example in Figure \ref{fig:case} to further demonstrate the importance of such details. From the similarity matrix, we can see that \method aligns both modalities quite well for all words but ConST struggles on words ``that'' and ``evolve'' as highlighted in blue boxes. This directly results in two translation errors of ConST. First, it fails to recover the clause structure implied by ``that''. Second, it omits ``evolve'' entirely in the translation. Though ConST still provides higher sentence similarity than \method, it fails to understand the subtlety inside the sentence. More examples are in Figure~\ref{fig:case_apdx}.

\begin{table}[t]
    \centering
    \begin{tabular}{l|ccc} 
        \toprule
         & \textbf{CTC} & \textbf{ConST} & \textbf{\method} \\
        \midrule
        Word-Level Sim & 0.08 & 0.44 & \textbf{0.65} \\
        Sentence-Level Sim & 0.13 & \textbf{0.34} & 0.30 \\
        \bottomrule
    \end{tabular}
    \caption{Average cosine similarity between word-level and sentence-level representations from speech and transcript. ASR and ST data are fixed to 10h and 1h respectively. \method achieves much higher word-level similarity than ConST and CTC. Though not optimized for sentence-level objective, \method still reaches close sentence-level similarity with ConST.}
    \label{tab:avg_cos}
\end{table}





\subsection{Why is \method better than CTC?}
\label{sec:ctc_bpe}



\begin{table}[t]
    \centering
    \begin{tabular}{l|cc}\toprule
        \textbf{Method} & 100h ASR & 370h ASR \\\midrule
        CTC & 18.3 & 20.4 \\
        CTC\textsubscript{Phoneme} & 14.3 & 19.0 \\
         \midrule
        \method & \textbf{21.0} & \textbf{22.7} \\
        \bottomrule
    \end{tabular}
    \caption{Case sensitive detokenized BLEU scores on MuST-C En-De tst-COMMON of CTC models with BPE (CTC by default) and with phoneme tokenization~(CTC\textsubscript{Phoneme}) and \method. 
    Fine-tuning ST data is fixed at 10h, while ASR data for pre-training is varied. The same MT pre-trained model is applied. 
    }
    \label{tab:ctc_phone}
\end{table}

\method treats words as base units to preserve acoustic boundaries and leverage knowledge of the pre-trained MT model, while CTC cannot achieve both merits simultaneously. CTC cannot benefit from word tokenization due to its extremely large vocabulary. To preserve acoustic boundaries, CTC requires phoneme or character tokenization. To leverage pre-trained MT model, CTC requires the same tokenization with MT model, i.e., BPE tokenization, but it has no guarantee on well-behaved acoustic boundaries.




\begin{figure}[t]
    \centering
    
    \begin{subfigure}{\linewidth}
        \centering
        \includegraphics[width=\textwidth]{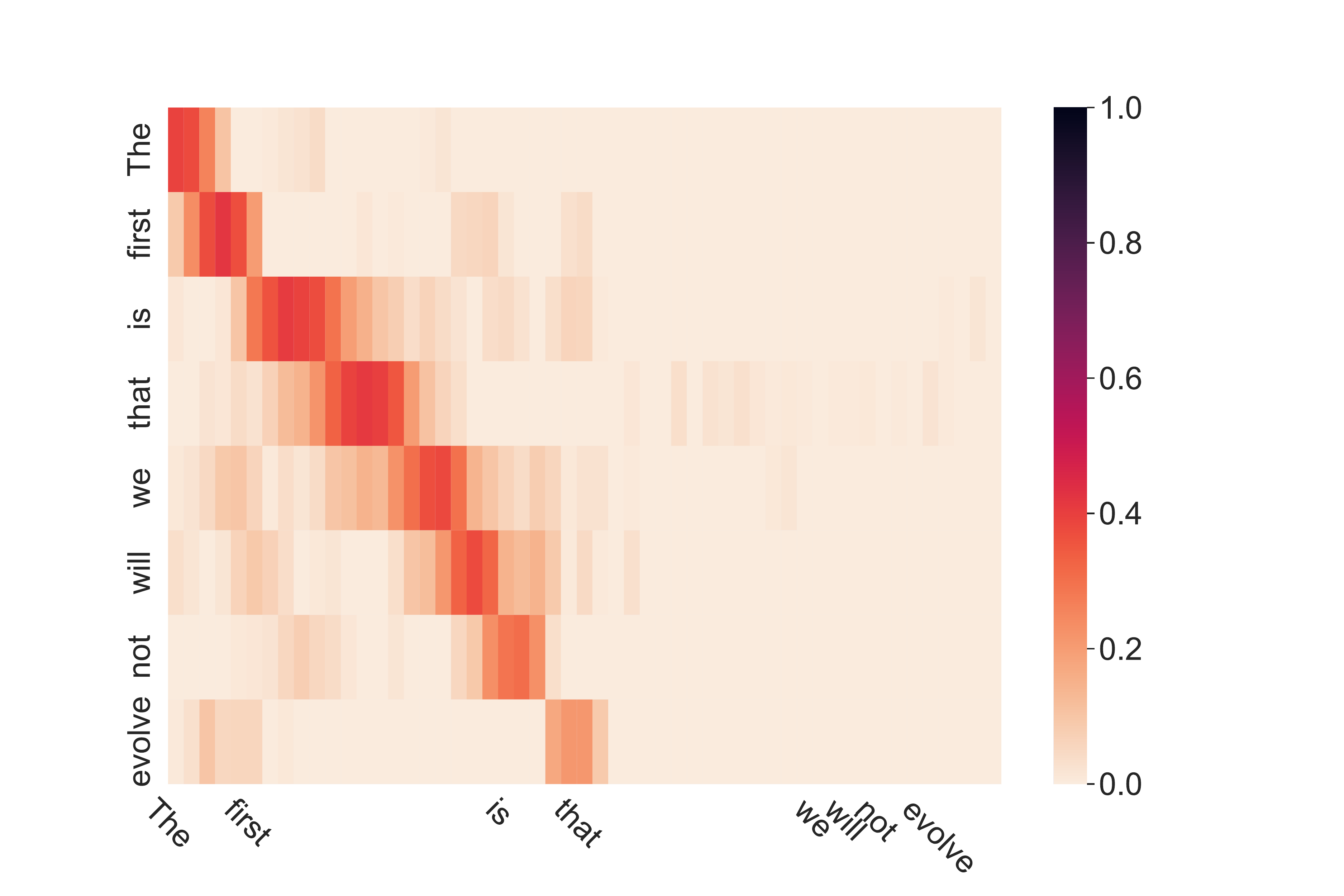}
        \caption{CTC}
        \label{fig:bad_align}
    \end{subfigure}
    
    \begin{subfigure}{\linewidth}
        \centering
        \includegraphics[width=\textwidth]{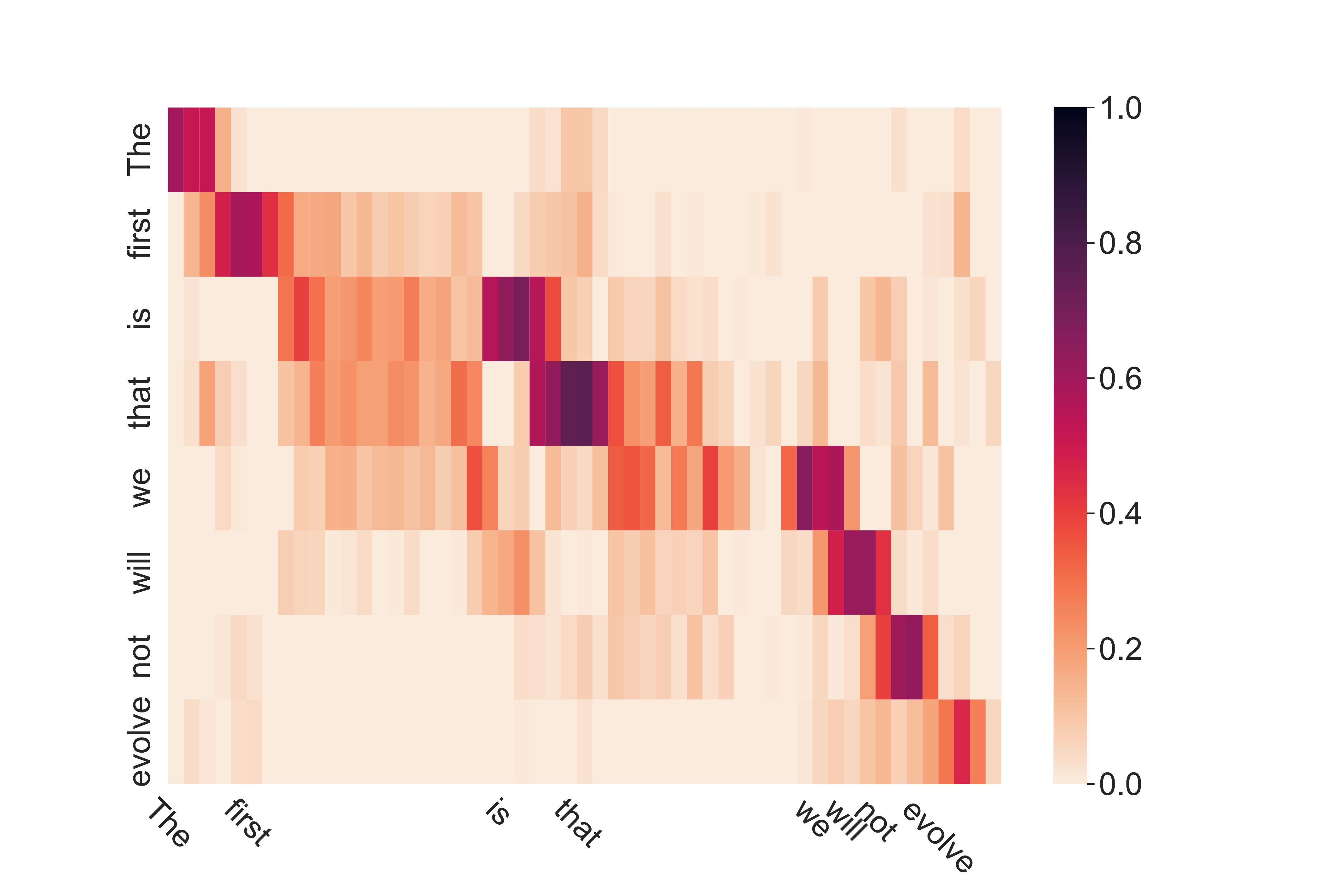}
        \caption{\method}
        \label{fig:good_align}
    \end{subfigure}
    
    \caption{Token-to-Frame embedding alignment matrix produced by models trained with CTC and \method respectively. Each row corresponds to a word and each column stands for a frame. Words in X-axis are placed according to their timestamps in speech to show how well the alignments are.}
    \label{fig:align}
\end{figure}

Figure~\ref{fig:align} gives a typical example of the token-to-frame embedding alignment matrix, indicating that \method learns more accurate alignment compared to CTC with BPE tokenization. 
Table~\ref{tab:avg_cos} also substantiates this quantitatively (0.65 and 0.30 \textit{v.s.} 0.08 and 0.13).

For acoustic tokenizations, we evaluate phoneme tokenization. We use the same phoneme vocabulary and grapheme-to-phoneme package as in \cite{tang-etal-2022-unified}. As shown in Table~\ref{tab:ctc_phone}, \method consistently outperforms CTC with phoneme tokenizations, proving its effectiveness in exploiting pre-trained MT knowledge.



\input{table/seqkd_cascade}

\subsection{Comparison with Sequence-Level Knowledge Distillation and Cascade ST}

One might argue that for the low-resource scenario we described above, sequence-level knowledge distillation (SeqKD, i.e. forward translation)~\cite{kim2016sequence} and building a cascaded system are the most straightforward way to leverage additional data and improve performance. 
The former trains an E2E model by expanding the ASR data translated from the existing MT model, while the latter builds the ASR and MT models separately. 
We compare \method and these two methods in various settings to demonstrate their differences. The implementation details of SeqKD and Cascade ST are described in Appendix \ref{sec:ft_cst}.

We show the result with 10h ASR and 1h ST data in Table \ref{tab:ft_cst}. Results of more data configurations are shown in Table \ref{tab:ft_cst_full} in the Appendix. We vary the amount of other pre-training resources: speech-only data and MT data. Speech-only data is used to pre-train speech encoder (i.e., wav2vec 2.0) and MT data is used to pre-train text embedding and joint Transformer in \method and MT models in SeqKD and Cascade ST.

All models drops 1$\sim$2 BLEU points when using 10\% MT data. Cascade ST performs better than \method when initialized with wav2vec 2.0. However, \method outperforms the cascade system when we train speech encoders from scratch. The ASR model trained from scratch in Cascade ST fails to produce meaningful transcript due to severe overfitting. Surprisingly, both \method and SeqKD are not sensitive to wav2vec 2.0 initialization. \method and SeqKD are also complementary to each other, and combining them leads to the highest BLEU score.

%% file: table/seqkd_cascade.tex
\begin{table}[t]
    \setlength{\belowcaptionskip}{-0.5cm}
    \centering
    \begin{tabular}{l|ccc}\toprule
        \textbf{MT Data}     & \textbf{4.6M}  & \textbf{4.6M} & \textbf{0.46M} \\
        \textbf{W2V2 Init}  & \ding{51} & \ding{55} & \ding{51} \\\midrule
        Cascade ST          & 15.6  & <5  & 14.1      \\
        SeqKD & 18.9  & 18.3  & 16.9      \\
        \midrule
        \method             & 14.1  & 14.3  & 12.5      \\
        \quad w/ SeqKD    & \textbf{19.5} & \textbf{19.2}  & \textbf{17.5}  \\\bottomrule
    \end{tabular}

    \caption{Case-sensitive detokenized BLEU scores on MuST-C En-De \texttt{tst-COMMON} set of \method, Cascade ST and Forward Translation. We vary the amount of MT training data from 4.6M to 0.46M and initializes the speech encoder with wav2vec 2.0 or random weights.}
    \label{tab:ft_cst}
\end{table}

%% file: 060conclusion.tex
In this work, we propose \method to align word-level speech and text embeddings. Experiments demonstrate the effectiveness of our method in both low-resource and regular ST settings. Analysis shows that our method can achieve better speech-text alignment which correlates well with its ST performance. 

%% file: 070limitation.tex
There are two main limitations in this work. 


First, instead of best ST performance given full data, our cross-modal pre-training only aims to demonstrate the effectiveness of our method in the low-resource ST setting. We realize that unified pre-training for both speech and text gradually becomes a dominant paradigm for ST and our future work is to fuse \method into a joint pre-training framework.

Second, we note that \citet{tang-etal-2022-unified} explores the possibility of pre-training MT models with phoneme tokenizations, though it is unclear if the phoneme-based MT model has an advantage over the BPE-based MT model. We follow the tradition of using the latter one and leave the comparison of them in future works.




%% file: 080impact.tex
\method has the potential to benefit speakers of low-resource languages. For example, their published video or speech can be better translated into other languages, so more viewers in the world can understand them, enabling deeper communication between different cultures. Though \method may be beneficial to cross-language communication, we do not encourage users to treat the translation generated by the E2E ST model as fully correct since they are far from perfect in practice. 


%% file: 090appendix.tex
\subsection{Statistics of Datasets}
\label{sec:data_stat}

We show statistics of MuST-C, IWSLT Mt-En, LibriSpeech and WMT datasets in Table \ref{tab:stat_mustc},\ref{tab:fewshot_stat},\ref{tab:stat_ls} and \ref{tab:stat_wmt}.

\begin{table}[h]
    \centering
    \begin{tabular}{l|cc}\toprule
        Direction & Hours & \# Sentence \\\midrule
        En-De     & 408   & 234K   \\
        En-Fr     & 492   & 280K   \\
        En-Es     & 504   & 270K   \\\bottomrule
    \end{tabular}
    \caption{Statistics of MuST-C.}
    \label{tab:stat_mustc}
\end{table}

\begin{table}[h]
    \centering
    \begin{tabular}{l|cc}\toprule
        Type                    & Hours & \# Sentence \\\midrule
        \multirow[l]{2}*{En-De ST}    & 1     & 0.6K      \\
                                & 10    & 5.8K      \\ \midrule
        \multirow[l]{4}*{En ASR}   & 10    & 5.8K      \\
                                & 100   & 58K       \\
                                & 370   & 216K      \\
                                & 1330  & 497K  \\\midrule
        Mt-En ST & 1 & 0.9K \\\midrule
        Mt ASR   & 10 & 6.7K
                                \\\bottomrule
    \end{tabular}
    \caption{Statistics of ST and ASR subsets in MuST-C En-De Low-Resource and IWSLT Mt-En.}
    \label{tab:fewshot_stat}
\end{table}

\begin{table}[H]
    \centering
    \begin{tabular}{l|ccc}\toprule
        Language & Hours & \# Sentence & \# Speaker \\\midrule
        En       & 960   &  281K       & 2338  \\\bottomrule
    \end{tabular}
    \caption{Statistics of LibriSpeech.}
    \label{tab:stat_ls}
\end{table}

\begin{table}[H]
    \centering
    \begin{tabular}{l|cc}\toprule
        Direction & Name & \# Sentence \\\midrule
        En-De     & WMT16   & 4.6M   \\
        En-Fr     & WMT14   & 40.8M   \\
        En-Es     & WMT13   & 15.2M   \\\bottomrule
    \end{tabular}
    \caption{Statistics of WMT.}
    \label{tab:stat_wmt}
\end{table}

\input{table/seqkd_cascade_2}
\input{table/wer}

\subsection{Impact of Hyperparameter}
\label{sec:ablation}

\begin{figure}[htbp]
    \centering
    \includegraphics[width=0.8\linewidth]{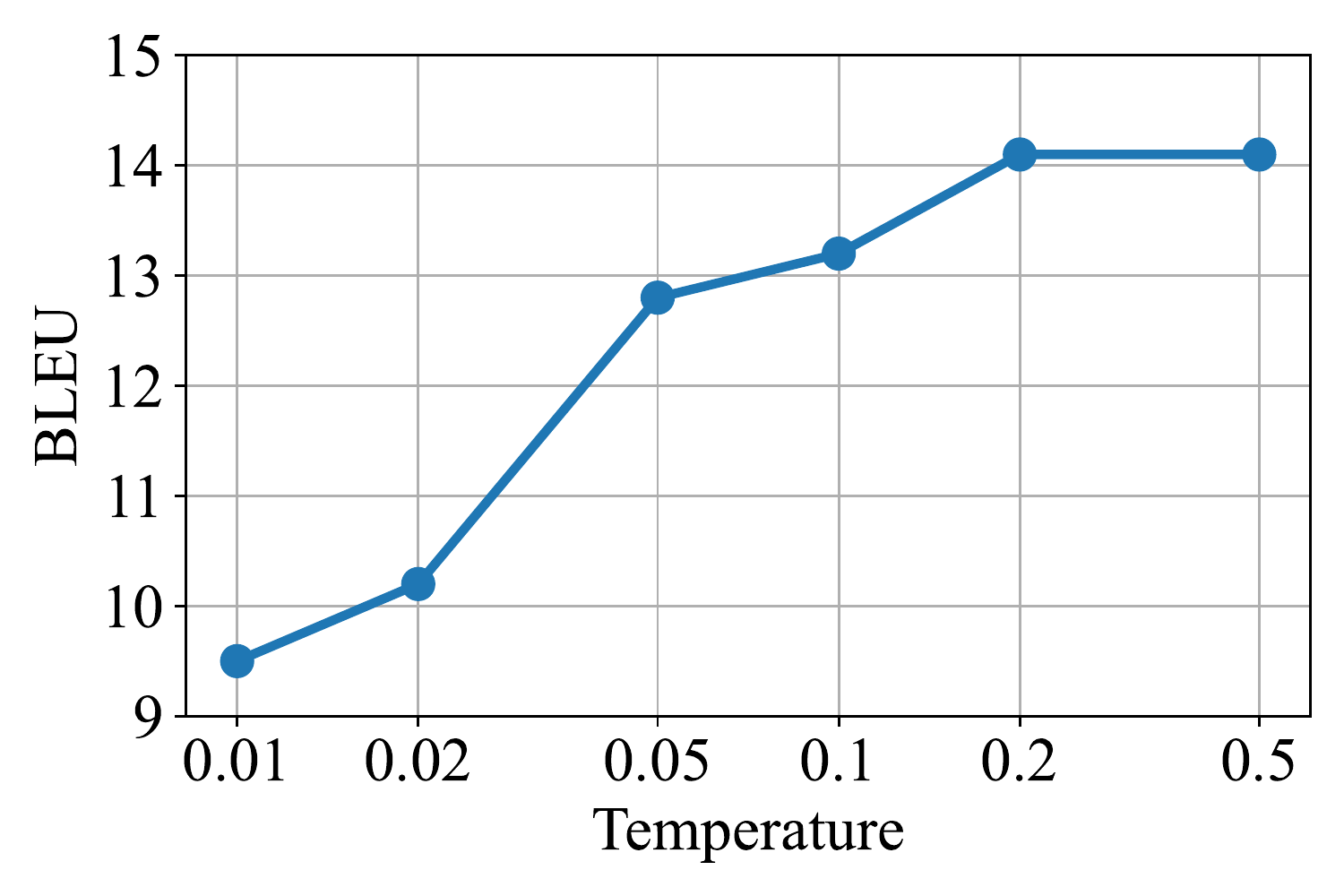}
    \caption{Case-sensitive detokenized BLEU scores on MuST-C En-De \texttt{tst-COMMON} set with different temperature values.}
    \label{fig:ablation_temp}
\end{figure}

\paragraph{Temperature} Figure \ref{fig:ablation_temp} demonstrates BLEU scores produced by different temperature values in low-resource ST setting with 10h ASR and 1h ST data. Higher temperature in general leads to higher BLEU scores, but the marginal improvement becomes negligible when $\tau>0.5$. 

\paragraph{Pooling and Layer} Here we compare different pooling mechanisms used to aggregate word-level representation and also different layers used to extract representations (before or after joint Transformer encoder). Table \ref{tab:pool_layer} shows BLEU scores in low-resource ST setting with 10h ASR and 1h ST data. \method is not very sensitive to pooling method, but sensitive to the layer selection. Aligning representations extracted before joint encoder delivers much better performance.

\begin{table}[h]
    \centering
    \begin{tabular}{c|cccc}\toprule
        Pooling & Mean & Max & Sum & Mean \\
        Layer   & Before & Before & Before & After \\\midrule
        BLEU    & 14.1 & 14.7 & 14.1 & 8.4  \\\bottomrule   
    \end{tabular}
    \caption{Case-sensitive detokenized BLEU scores on MuST-C En-De \texttt{tst-COMMON} set with different pooling mechanisms (used to aggregate word-level feature) and layers (used to extract speech and text representations).}
    \label{tab:pool_layer}
\end{table}

\subsection{More Examples of \method versus ConST}

\begin{figure*}[ht]
    \centering
    
    \begin{subfigure}{\textwidth}
        \centering
        \includegraphics[width=\textwidth]{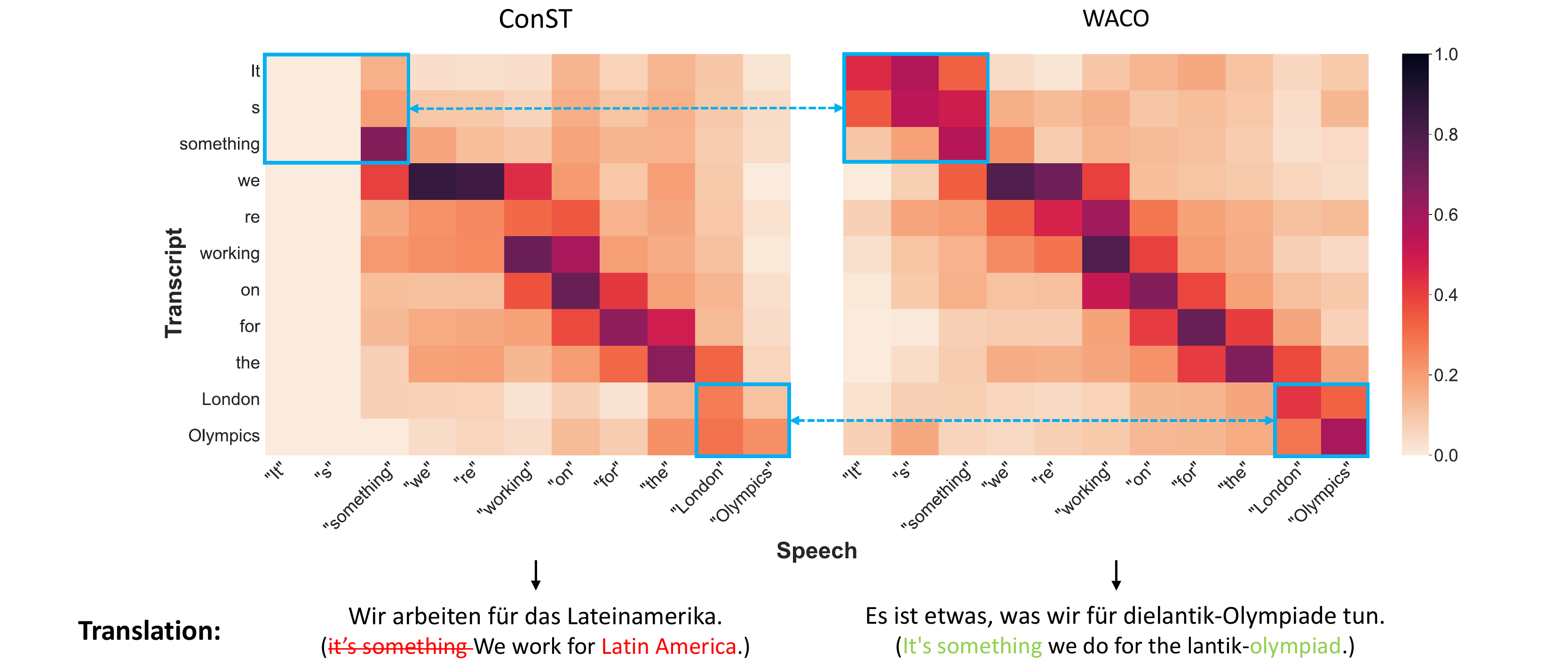}
    \end{subfigure}
    
    \vspace{+1cm}
    
    \begin{subfigure}{\textwidth}
        \centering
        \includegraphics[width=\textwidth]{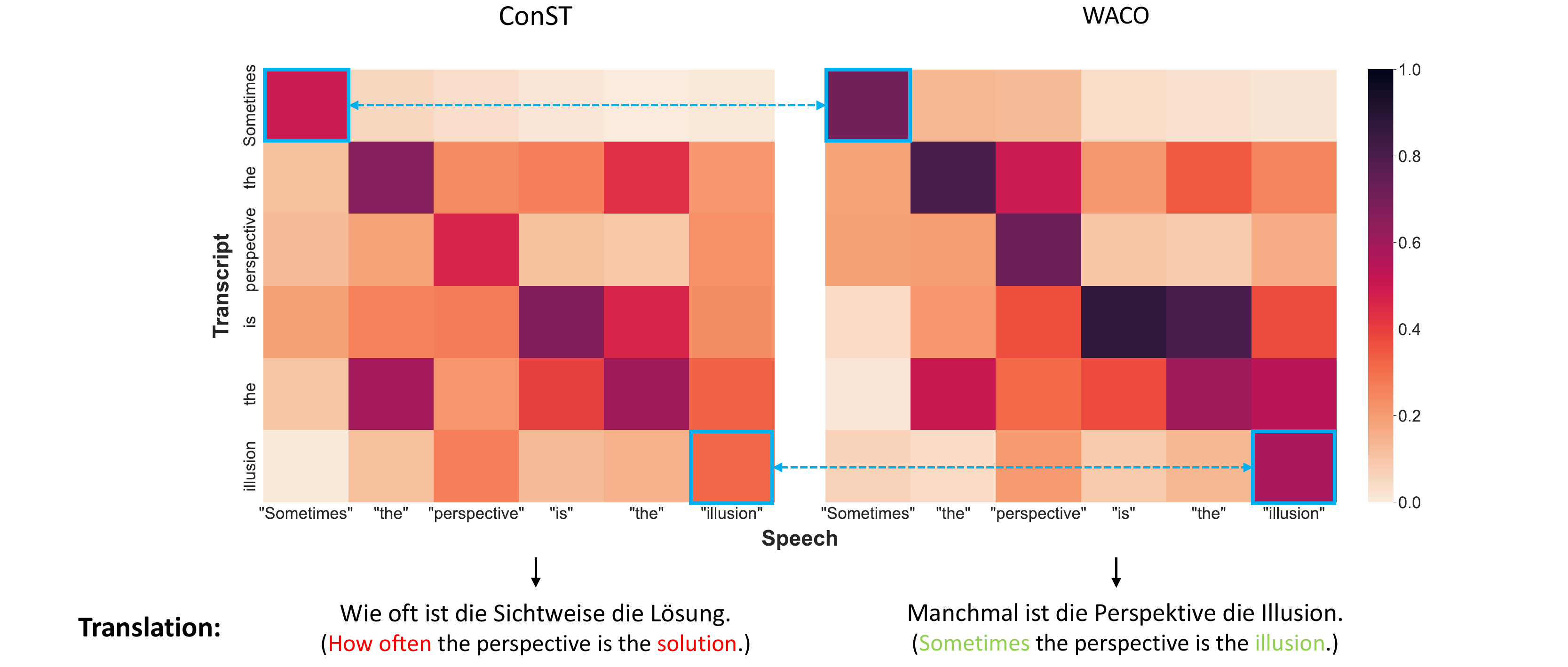}
    \end{subfigure}

    \caption{Two additional examples with speech-text alignment matrices and translations of \method and ConST.}
    \label{fig:case_apdx}
\end{figure*}

We show two more examples that \method achieves more accurate translation than ConST by better speech-text alignment in Figure \ref{fig:case_apdx}.

\subsection{Loss Curves for Cross-Modal Pre-training}

\begin{figure}[ht]
    \centering
    
    \begin{subfigure}{\linewidth}
        \centering
        \includegraphics[width=\linewidth]{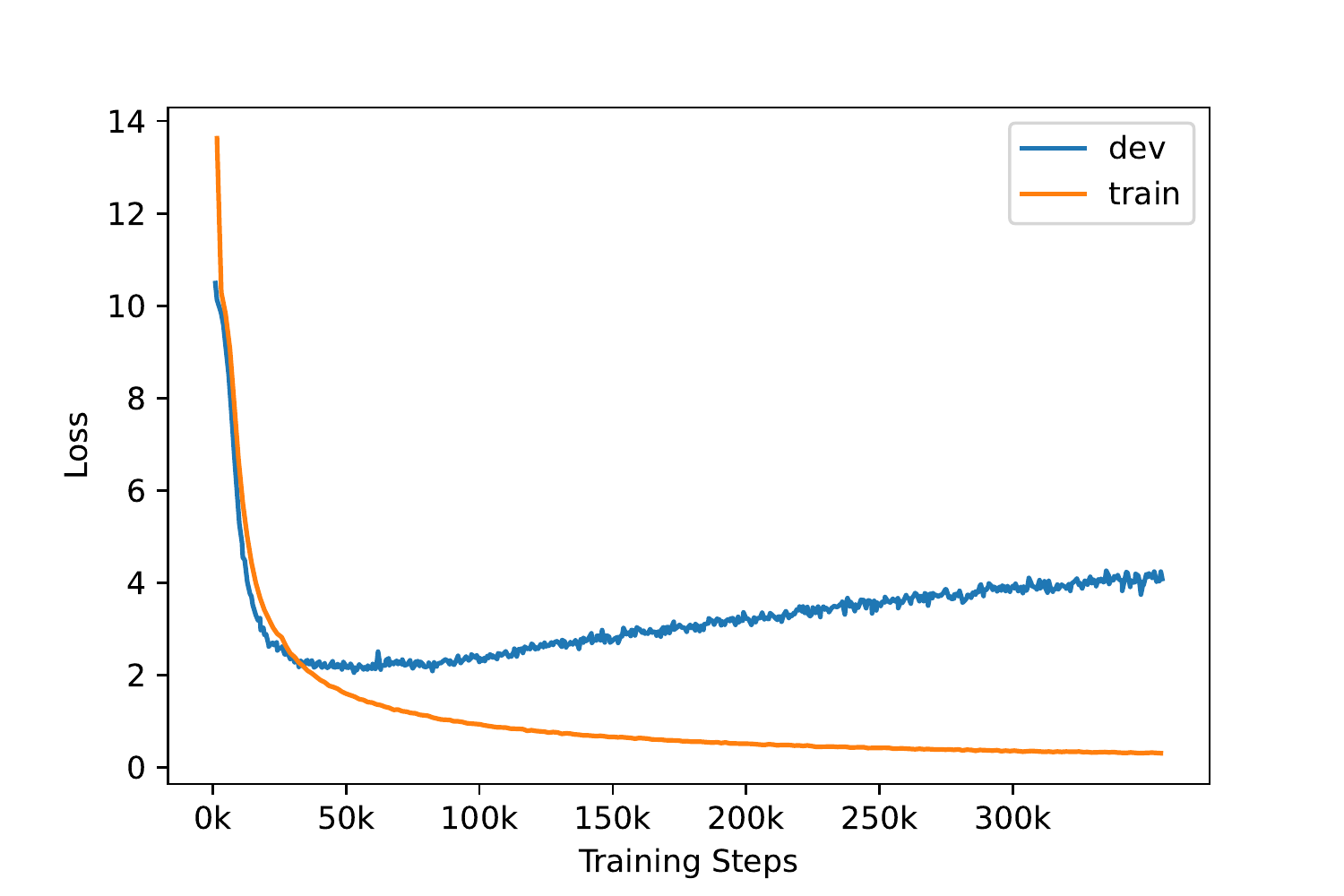}
        \caption{CTC with BPE tokenization}
        \label{fig:ctc_loss}
    \end{subfigure}
    
    \begin{subfigure}{\linewidth}
        \centering
        \includegraphics[width=\linewidth]{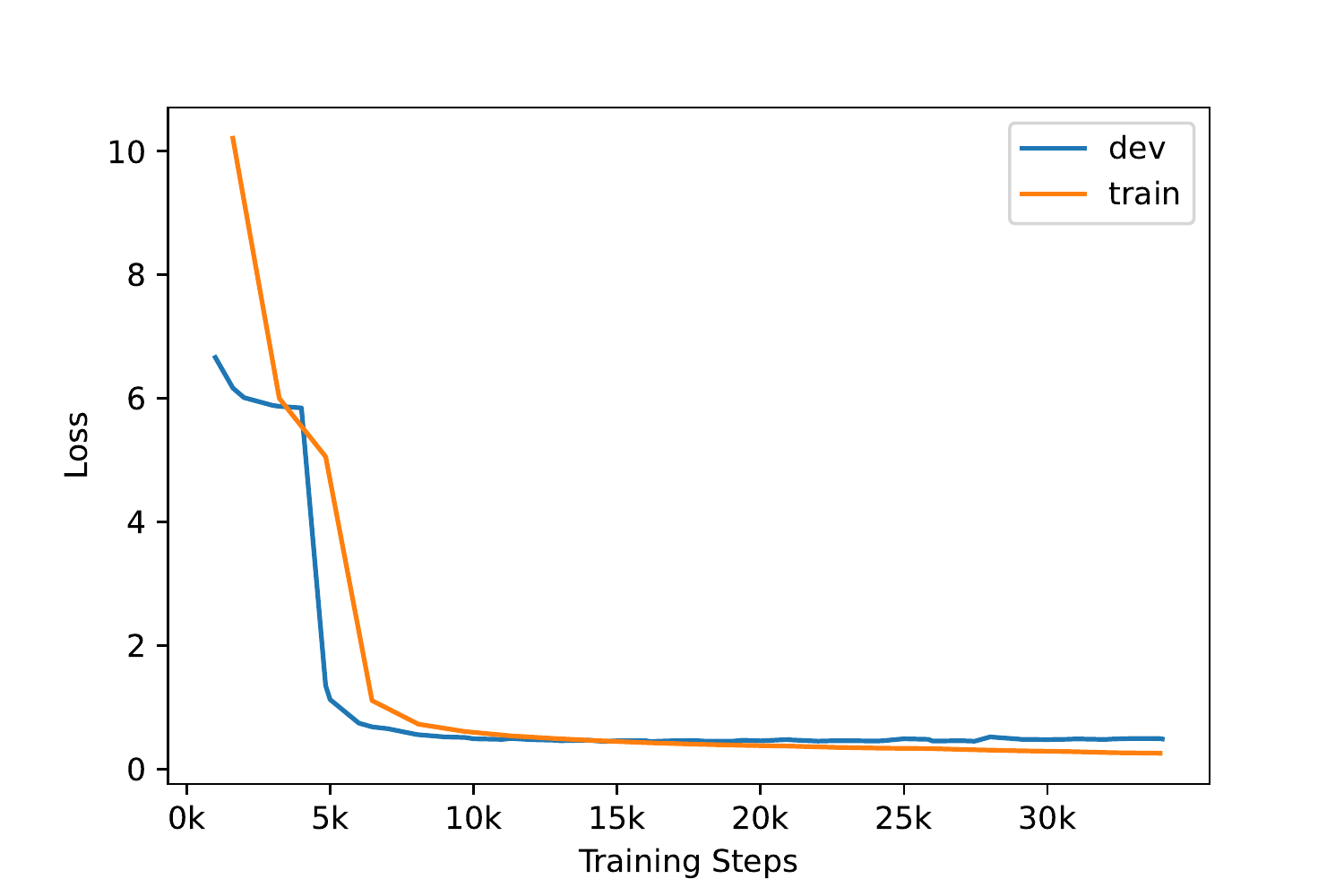}
        \caption{CTC with phoneme tokenization}
    \end{subfigure}
    
    \begin{subfigure}{\linewidth}
        \centering
        \includegraphics[width=\linewidth]{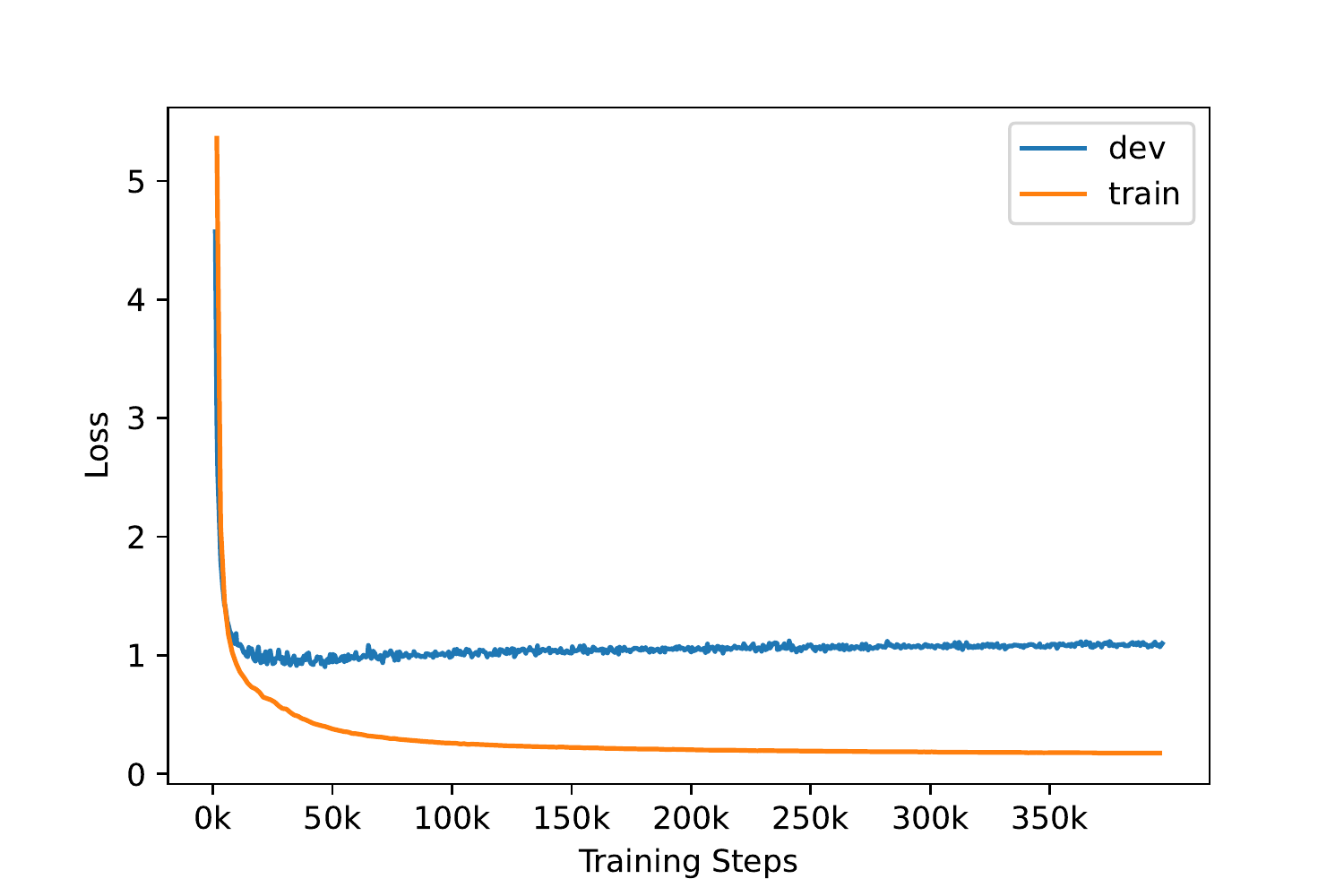}
        \caption{WACO}
    \end{subfigure}
    
    \caption{Loss curves of various cross-modal pre-training method. CTC with BPE tokenization cannot generalize well to unseen speech (cannot reach below 2 on dev set).}
    \label{fig:loss_curve}
\end{figure}

We present pre-training loss curves of CTC with both BPE and phoneme tokenizations, and \method in Figure \ref{fig:loss_curve}.

\subsection{MT Pre-training}
\label{sec:mt_pt}

We use the same vocabulary and SentencePiece model described in Section~\ref{sec:exp_setup} to tokenize the WMT datasets. The model is optimized with Adam. The learning rate starts at 1e-7, warmed up to 7e-4 by 4k steps and then decays following the inverse square root schedule with a minimum learning rate of 1e-9. The maximum number of tokens in a batch is 8192. We select the checkpoint with the highest BLEU (beam size 4, length penalty 0.6) on the WMT validation set.

\subsection{Sequence-Level Knowledge Distillation and Cascade ST}
\label{sec:ft_cst}

\paragraph{SeqKD} We apply the same MT model used to initialize joint Transformer in \method to translate the transcript of ASR data into target language, which is German in our case. The translation is produced by beam search with width 10 and length penalty 0.6. After we construct the pseudo ST dataset (10h), we combine it with the real-world ST dataset (1h) and obtain an 11h triplet ST dataset. Then we follow the same fine-tuning procedure in low-resource ST setting, i.e., cross-entropy losses, to obtain the final ST model.

\paragraph{Cascade ST} We follow the same fine-tuning procedure used in \cite{baevski2020wav2vec} to fine-tune wav2vec 2.0 small on 10h ASR data to obtain the ASR model. The final transcript is decoded by Viterbi algorithm. We use the same MT model and configuration as in SeqKD to translate the English transcript into German translation.

\subsection{Results of More Data Configurations}

We compare \method and Cascade ST with or without wav2vec2 in more data configurations as in Table \ref{tab:main_table} except for 1330-hour ASR. The results are shown in Table \ref{tab:ft_cst_full}.

\subsection{Word Error Rate of using \method as an ASR model} 

The word error rates (WER) of \method with different data configurations are shown in Table \ref{tab:wer}.

\subsection{Training Efficiency}
\label{sec:efficiency}
The computation cost of calculating \method loss function is higher than that of sentence-level methods (e.g., ConST) and in our profiling results, it takes 20$\sim$30\% time of a forward pass. However, \method converges much faster, in terms of number of iterations, than ConST and CTC. In the 100h ASR data case, WACO only needs <25k iterations to converge while both ConST and CTC requires >50k iterations. This makes WACO more time efficient overall.

%% file: table/seqkd_cascade_2.tex
\begin{table*}[t]
    \centering

    \begin{tabular}{l|ccccc}\toprule
        \textbf{ST Data} &  \multicolumn{3}{c}{1h} & \multicolumn{2}{c}{10h} \\
        \cmidrule(lr){2-4} \cmidrule(lr){5-6} 
        \textbf{ASR Data} & 10h & 100h & 370h & 100h & 370h \\
        \midrule
        Cascade w/o w2v2 & <1 & 4.7 & 10.7 & 4.7 & 10.7 \\
        WACO w/o w2v2  &   \textbf{14.3} & \textbf{15.4} & \textbf{15.3} & \textbf{20.9} & \textbf{22.3} \\
        \midrule 
        Cascade w/ w2v2 & \textbf{15.6} & \textbf{17.2} & \textbf{18.0} & 17.2 & 18.0 \\
        WACO w/ w2v2  &   14.1 & 16.2 & 16.6 & \textbf{21.0} & \textbf{22.7} \\
        \bottomrule
    \end{tabular}
    
    \caption{Case-sensitive detokenized BLEU scores on MuST-C En-De \texttt{tst-COMMON} set of \method and Cascade ST. We initializes the speech encoder with wav2vec 2.0 or random weights instead. }
    \label{tab:ft_cst_full}
\end{table*}

%% file: table/wer.tex
\begin{table*}[t]
    \centering

    \begin{tabular}{l|ccccccc}\toprule
        \textbf{ST Data} &  \multicolumn{4}{c}{1h} & \multicolumn{3}{c}{10h} \\
        \cmidrule(lr){2-5} \cmidrule(lr){6-8} 
        \textbf{ASR Data} & 10h & 100h & 370h & 1330h &  100h & 370h & 1330h \\
        \midrule
        \method & 45.5 & 40.6 & 42.7 & 33.3 & 24.9 & 21.7 & 21.3 \\
        \bottomrule
    \end{tabular}
    
    \caption{Word Error Rate (WER) of \method on ASR part of MuST-C En-De \texttt{tst-COMMON} set.}
    \label{tab:wer}
\end{table*}